\documentclass[conference, 10pt]{IEEEtran}
\IEEEoverridecommandlockouts
\usepackage{cite}
\usepackage{amsmath,amssymb,amsfonts}
\usepackage{graphicx}
\usepackage{textcomp}
\usepackage{xcolor}
\usepackage{url}
\usepackage{placeins}
\usepackage{float}
\usepackage{adjustbox}
\usepackage{tabularx,colortbl}
\usepackage{ifthen}
\usepackage{caption,subcaption}
\usepackage{amsmath}
\usepackage{booktabs}
\renewcommand{\thetable}{\Roman{table}}
\usepackage{hyperref}
\usepackage{cleveref}
\let\labelindent\relax
\usepackage{enumitem}
\usepackage{stfloats}
\usepackage{wrapfig}
\usepackage{verbatim}
\usepackage{booktabs}
\usepackage{flushend}
\usepackage{amsmath,amssymb,amsfonts}
\usepackage{mathtools} 
\usepackage{multirow}
\usepackage{algorithm}
\usepackage{algpseudocode}
\usepackage{pifont}

\newcommand{\cmark}{\ding{51}} 
\newcommand{\xmark}{\ding{55}} 

\def\BibTeX{{\rm B\kern-.05em{\sc i\kern-.025em b}\kern-.08em
    T\kern-.1667em\lower.7ex\hbox{E}\kern-.125emX}}

\begin{document}
\newcolumntype{P}[1]{>{\centering\arraybackslash}p{#1}}
\newcolumntype{M}[1]{>{\centering\arraybackslash}m{#1}}
\setlength{\textfloatsep}{10pt plus 1.0pt minus 2.0pt}
\setlength{\dbltextfloatsep}{10pt plus 1.0pt minus 2.0pt}
\setlength{\floatsep}{5pt plus 1.0pt minus 2.0pt}
\setlength{\dblfloatsep}{5pt plus 1.0pt minus 2.0pt}
\setlength{\intextsep}{10pt plus 1.0pt minus 2.0pt}

\title{\LARGE \bf Enhancing 3D Robotic Vision Robustness by Minimizing Adversarial Mutual Information through a Curriculum Training Approach}

\author{Nastaran Darabi, Dinithi Jayasuriya, Devashri Naik, Theja Tulabandhula, and Amit Ranjan Trivedi
\thanks{Authors are with the University of Illinois at Chicago (UIC). This work was supported in part by COGNISENSE, one of seven centers in JUMP 2.0, a Semiconductor Research Corporation (SRC) program sponsored by DARPA, and NSF funding \#2235207. Corresponding Authors Email: {\tt\small ndarab2@uic.edu, amitrt@uic.edu}}
}
\maketitle

\begin{abstract} 
Adversarial attacks exploit vulnerabilities in a model’s decision boundaries through small, carefully crafted perturbations that lead to significant mispredictions. In 3D vision, the high dimensionality and sparsity of data greatly expand the attack surface, making 3D vision particularly vulnerable for safety-critical robotics. To enhance 3D vision's adversarial robustness, we propose a training objective that simultaneously minimizes prediction loss and mutual information (MI) under adversarial perturbations to contain the upper bound of misprediction errors. This approach simplifies handling adversarial examples compared to conventional methods, which require explicit searching and training on adversarial samples. However, minimizing prediction loss conflicts with minimizing MI, leading to reduced robustness and catastrophic forgetting. To address this, we integrate curriculum advisors in the training setup that gradually introduce adversarial objectives to balance training and prevent models from being overwhelmed by difficult cases early in the process. The advisors also enhance robustness by encouraging training on diverse MI examples through entropy regularizers. We evaluated our method on ModelNet40 and KITTI using PointNet, DGCNN, SECOND, and PointTransformers, achieving 2--5\% accuracy gains on ModelNet40 and a 5--10\% mAP improvement in object detection. Our code is publicly available at \hyperlink{https://github.com/nstrndrbi/Mine-N-Learn}{\textcolor{magenta}{https://github.com/nstrndrbi/Mine-N-Learn}}.
\end{abstract}

\section{Introduction}
Adversarial attacks on machine learning systems manipulate model outputs by introducing subtle perturbations that lead to significant misclassifications. Common attacks such as Projected Gradient Descent (PGD) \cite{zhang2019theoretically}, Fast Gradient Sign Method (FGSM) \cite{liu2019extending}, and Carlini-Wagner (CW) \cite{carlini2017towards} exploit vulnerabilities in a model's decision boundaries by identifying regions where small, carefully crafted perturbations can cause significant misclassifications \cite{Xiang_2019_CVPR}. Such attacks are particularly concerning for safety-critical applications, such as autonomous driving, where, for example, misclassifying a stop sign could prevent the vehicle from braking \cite{szegedy2013intriguing}.

Adversarial training \cite{kurakin2016adversarial} has emerged as a key defense strategy against such attacks, where the model is trained on both clean and adversarial examples to improve its robustness. Among prior studies, Goodfellow et al. \cite{goodfellow2014explaining} incorporated adversarial perturbations during training to reshape decision boundaries and reduce vulnerability. Madry et al. \cite{madry2017towards} extended this with a min-max formulation, optimizing the model for worst-case adversarial examples, thereby significantly improving resistance to strong attacks like Projected Gradient Descent (PGD). Zhang et al. \cite{zhang2019theoretically} proposed TRADES, which balances robustness and accuracy by explicitly managing the trade-off between clean and adversarial performance. Wong et al. \cite{wong2020fast} introduced fast training to reduce computational costs.

While adversarial attacks and their defenses have been widely explored in 2D images, they present distinct challenges for 3D vision--LiDAR and radar. With the increasing use of 3D sensors in mission-critical robotics for tasks such as depth perception \cite{wang2023visual}, object detection \cite{hasan2022lidar, stutts2024mutual}, and environmental mapping \cite{shukla2021ultralow}, the high dimensionality and sparsity of these outputs expands the attack surface and increases potential adversarial examples exponentially. This vast adversarial space complicates training robust models, as excessive adversarial data can degrade accuracy on the clean data, leading to over-fitting on adversarial examples and poor generalization.

\begin{figure}
    \centering
    \includegraphics[width=0.9\linewidth]{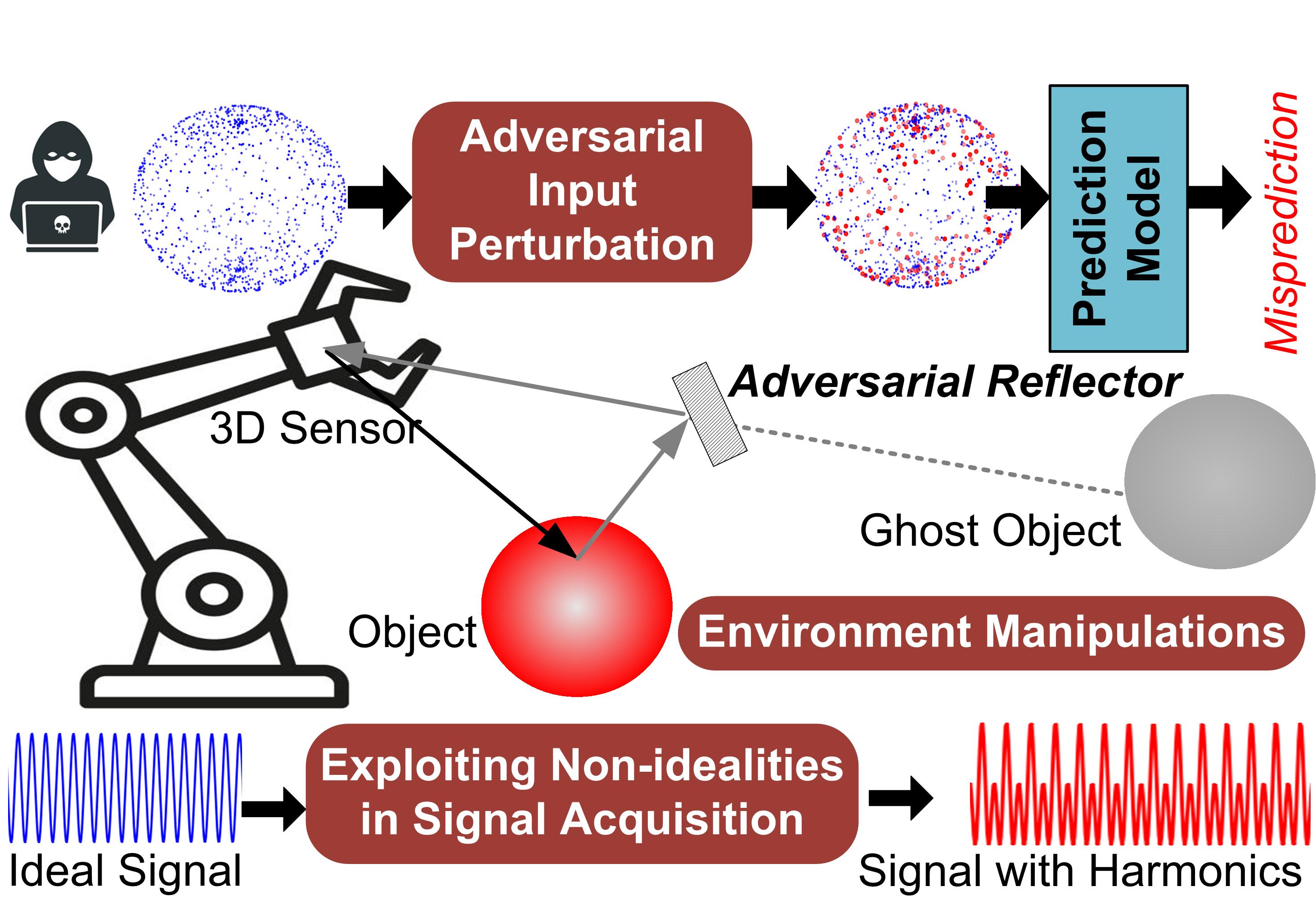}
    \caption{3D vision sensors used in safety-critical robotics are vulnerable to various adversarial attacks. For instance, sensor data can be manipulated by adding or removing points, or environments can be altered by placing reflective surfaces that create ghost objects. Non-idealities of 3D sensor circuits can also be exploited such as by leveraging their harmonics to create false objects in the sensor’s field of view.}
    \label{fig:motivation}
\end{figure}

Addressing the challenges of robust 3D robotic vision for safety-critical applications, we demonstrate that a training objective minimizing both prediction loss and the mutual information (MI) of outputs against adversarial perturbations significantly enhances robustness. These ``corner cases,'' where models struggle the most to generalize due to minimal predictive information and weak, non-redundant features, are especially vulnerable to adversarial attacks. By selectively targeting such cases, we efficiently improve adversarial robustness while maintaining high accuracy on clean data. Additionally, we show that simultaneously minimizing MI and prediction loss under constrained model capacity can lead to catastrophic forgetting due to conflicting objectives. To address this, we integrate curriculum training, with curriculum parameters such as pacing functions extracted from the training data. Our key contributions are:
\begin{itemize}[itemsep=2pt,topsep=2pt,leftmargin=10pt]
\item We discuss that minimizing MI under adversarial perturbations reduces the upper bound on prediction errors for adversarial inputs. Leveraging this, we propose a training objective that simultaneously minimizes both prediction loss and MI under perturbations. This approach eliminates the need for explicitly searching adversarial examples, which is complex for 3D-vision datasets due to their high dimensionality, sparsity, and irregularity, thus making comprehensive adversarial training coverage challenging.
\item For model scalability, our loss function leverages neural networks for MI extraction. We also design data-driven curriculum advisors that guide the training process to incrementally enforce adversarial robustness constraints, thus effectively balancing the conflicting objectives of prediction accuracy and adversarial robustness. The advisors also efficiently explore the training space by encouraging high entropy of MI cases to comprehensively cover training space.
\item We rigorously evaluated the proposed framework on ModelNet40 \cite{wu20153d} and KITTI \cite{Geiger2012CVPR} datasets against eleven attack mechanisms, and across various network architectures including PointNet \cite{qi2017pointnet}, PointTransformer \cite{zhao2021point}, SECOND \cite{yan2018second}, PointPillars \cite{lang2019pointpillars}, and TED \cite{wu2023transformation}. Our defense mechanism improved detection accuracy (mAP) across the PointPillars, SECOND, and TED models, with increases ranging from 5--10\% under adversarial attacks. The ablation study further showed that combining adversarial training (AT), low-mutual information extraction (MINE), and data-driven curriculum training (CT) resulted in improvements ranging from 2.5-10\%, with TED's Pedestrian detection under IFGM attack \cite{liu2019extending} rising from 55.83\% to 66.27\%.
\end{itemize}

\section{Background}
\subsection{Adversarial Attacks on Learning-based 3D Vision}
Fig. \ref{fig:motivation} illustrates various practical adversarial attack mechanisms on 3D vision systems. At the top, the sensed input from a 3D vision sensor can be adversarially manipulated such that, despite imperceptible changes, the output fed into prediction models experiences a significant accuracy reduction. Among various prior works exploring such attacks, Hamdi et al. \cite{hamdi2020advpc} introduced AdvPC, by creating transferable adversarial perturbations on 3D point clouds, while Wen et al. \cite{wen2020geometry} proposed geometry-aware adversarial examples targeting structural properties of point clouds. These works underscore the vulnerability of 3D models to adversarial attacks. 

Moreover, the physical environment of 3D sensors can be exploited. For instance, in the middle of Fig. \ref{fig:motivation}, an adversary can place reflectors in the surroundings, causing multi-path reflections. As 3D sensors actively sense their environment, reflected signals can create the appearance of ghost objects due to these artifacts. Similar to the above software-based attacks, the adversarial surfaces can be designed to maximally interfere with the prediction loss, resulting in false positives or inaccuracies in detection that lead to failures \cite{hau2021shadow}. 

Likewise, non-linearities in 3D vision acquisition circuits lead to harmonic distortions and can be exploited to create ghost targets. E.g., when the radar signal passes through non-linear components, such as analog amplifiers or mixers, the system’s output \( y(t) \) can be expressed as a power series:
\(
y(t) = a_1 x(t) + a_2 x^2(t) + a_3 x^3(t) + \dots
\)
where \( x(t) \) is the input signal. For a radar signal \( r(t) = \frac{A}{R} \cos(2 \pi f_0 (t - \tau) + \phi) \), the second-order non-linearity generates a second harmonic:
\(
a_2 \frac{A^2}{2R^2} \left(1 + \cos(4 \pi f_0 (t - \tau) + 2 \phi)\right)
\)
These harmonics can create ghost targets due to range and velocity ambiguities. Adversaries may exploit this by injecting a jamming signal \( j(t) = B \cos(2 \pi f_j t + \psi) \), which generates harmonics, such as \( 2f_j \), that overlap with the radar’s operating frequency \( f_0 \), thus creating ghost targets:
\(
y_j(t) = a_2 B^2 \cos(4 \pi f_j t + 2 \psi)
\). The presence of ghost targets can be adversarially optimize to induce maximal deviations in the overall prediction loss.\footnote{We acknowledge our discussions with Alyosha Molnar, Cornell University, to leverage such analog hardware non-idealities in 3D vision sensors for crafting adversarial attacks.}

\subsection{3D Vision Point Cloud Adversarial Defenses}
Several defense strategies have also been developed to improve 3D models' robustness against adversarial attacks. Dong et al. \cite{dong2020self} proposed a self-robust framework leveraging gather-vector guidance, while Wu et al. \cite{wu2021adversarial} introduced virtual adversarial training for point cloud classification. Huang et al. \cite{huang2022shape} focused on shape-invariant adversarial training based on geometric transformations. Yan et al. \cite{yan2022survey} provided a comprehensive survey on 3D adversarial attacks and defenses, and Ji et al. \cite{ji2023benchmarking} offered a hybrid training method incorporating various adversarial examples. 

Unlike prior methods that focused on geometric transformations or hybrid augmentation, we pursue a training objective on simultaneous minimization of prediction loss as well as adversarial vulnerabilities. For this, our training loss function integrates adversarial robustness by minimizing MI of adversarial perturbations alongside prediction loss; ideal for 3D vision datasets, where covering the space of adversarial examples is challenging due to high input dimensionality.

\section{Generalized Framework for Adversarial Point Cloud Manipulations}
Point cloud adversarial attacks can be categorized into three main types: \textit{Point Addition Attacks}, \textit{Point Removal Attacks}, and \textit{Point Shifting Attacks}, each detailed below: 

\vspace{2pt}
\noindent \textbf{(i) Point Addition Attack:} In this attack, the attacker strategically places new points \( X_\text{add} \in \mathbb{R}^{k \times 3} \), such as using the $CW$ attack \cite{carlini2017towards}, to determine optimal positions that deceive the predictor. Other methods include adding clusters or adversarial objects \cite{Xiang_2019_CVPR, ji2023benchmarking}. The attack objective is:
\begin{equation}
    \min_{X_\text{add}} \left[ L(X \cup (X_\text{add})) + \lambda D(X, X \cup (X_\text{add})) \right],
\end{equation}
where \(  L() \) is the adversarial loss that measures the effectiveness of the attack, \( D(X, X \cup (X_{add})) \) is a distance function that ensures minimal perceptual deformation, and \( \lambda \) controls the trade-off between attack success and perceptibility. 

\vspace{2pt}
\noindent \textbf{(ii) Point Removal Attack:} Likewise, for point removal attacks, the attacker selectively removes points from the point cloud, making the object appear incomplete or altered to the classifier. The attack objective is:
\begin{equation}
    \min_{X_{\text{drop}}} \left[ L(X \setminus X_{\text{drop}}) + \lambda D(X, X \setminus X_{\text{drop}}) \right]
\end{equation}
where \(X \setminus X_{\text{drop}}\) is the point cloud after point removal, \(L()\) is the adversarial loss, \(D(X, X \setminus X_{\text{drop}})\) is a distance function, and \(\lambda\) trades-off attack success with perceptibility.

\vspace{2pt}
\noindent\textbf{(ii) Point Shifting Attacks:}
Many prior works have focused on shifting attacks that subtly alter points within a point cloud to distort geometry and confuse models. For instance, Iterative Fast Gradient Method (IFGM) extends FGSM by iterating to create refined adversarial examples \cite{liu2019extending}. Methods like Projected Gradient Descent (PGD) and Perturb attacks ensure perturbed points stay on the object’s surface or avoid outliers \cite{liu2019extending}. Advanced techniques like GeoA3 \cite{wen2020geometry} and SIA \cite{huang2022shape} add geometric awareness, while AdvPC \cite{hamdi2020advpc} uses autoencoders for better transferability. AOF \cite{liu2022boosting} introduces frequency-domain perturbations, targeting low-frequency components for robustness. The table below summarizes the objective functions/perturbations of these attacks. \(h_\theta\) is the neural network, \(y\) is the ground truth, and adversarial perturbations \( \rho \) are applied to a subset of points \( X_{\text{shift}} \), resulting in adversarial point cloud \( X' = X + \rho \), with \( \lambda \) balancing strength and perceptibility.

\begin{table}[ht!]
    \centering
    \setlength{\tabcolsep}{4pt} 
    \renewcommand{\arraystretch}{1.2} 
    \caption{Shifting Attacks and their Objective Functions/Perturbations}
    \begin{tabular}{p{1cm}|p{7cm}}
    \hline
    \textbf{Attack} & \textbf{Objective Function/Perturbation} \\ \hline
    IFGM & 
    $\rho = \lambda \cdot \text{sign}(\nabla_X L(X, y))$ \\     
    PGD & 
    $\rho_{t+1} = \text{Proj}_\epsilon \left(\rho_t + \alpha \cdot \text{sign}(\nabla_X L(X + \rho_t, y))\right)$ \\     
    Perturb & 
    $\min_{\rho} \left[L(h_\theta(X + \rho), y) + \lambda \sum_{i=1}^n \|\rho_i\|_2 \right]$ \\     
    KNN & 
    $\min_{\rho} \left[L(h_\theta(X + \rho), y) + \lambda \cdot \text{KNN}(X + \rho)\right]$ \\     
    GeoA3 & 
    $\min_{\rho} \left[L(h_\theta(X + \rho), y) + \lambda \sum_{i=1}^n C(X_i, \rho_i)\right]$ \\     
    L3A & 
    $\min_{\rho} \left[L(h_\theta(X + \rho), y) + \lambda \|\nabla_\rho L(X + \rho, y)\|_2\right]$ \\     
    AdvPC & 
    $\min_{\rho} \left[L(h_\theta(X + \rho), y) + \lambda \cdot D(AE(X), AE(X + \rho))\right]$ \\     
    SIA & 
    $\min_{\rho} \left[L(h_\theta(X + \rho), y) + \lambda \sum_{i=1}^n \|\rho_i\|_2\right], \quad \rho_i \text{ is tangent}$ \\     
    AOF & 
    $\min_{\rho} \left[L(h_\theta(X + \rho), y) + \lambda \cdot \text{LowFreq}(X + \rho)\right]$ \\ \hline
    \end{tabular}
    
    \label{tab:attacks}
\end{table}

\section{Minimizing Input-Output Mutual Information for Adversarial Robustness}
\subsection{Adversarial Risk \textit{vs.} Mutual Information}
Gradient-based adversarial attacks become especially effective in low input-output MI spaces, where model uncertainty is higher. In this work, we explore a targeted adversarial training on these vulnerable decision boundaries. Standard adversarial training aims to minimize:
\begin{equation}
\min_{\theta} \mathbb{E}_{(X, y) \sim D} \left[ \max_{\rho \in \mathcal{S}} \mathcal{L}(h_\theta(X'), y) \right],
\end{equation}
where \( \mathcal{L} \) is the loss function, \( \mathcal{S} \) is adversarial perturbation set, and \( D \) is the data distribution. \(X \) is the clean input data, \(\rho\) is the adversarial perturbation, and \(X' = X + \rho \) is the corresponding adversarial input data. \( h_\theta() \) is the prediction model. \(y \) is ground truth output and \(y' \) is model's prediction. 


Comparatively, in this work, we pursue a training objective that together minimizes both prediction loss as well as input-output adversarial MI by adding a loss term:
\begin{equation}
\min_{\theta} \mathbb{E}_{(X, y) \sim D+\rho} \left[ \mathcal{L}(h_\theta(X'), y)  + \lambda \cdot I(\rho; y')\right].
\end{equation}
Here, $D+\rho$ is an augmented dataset used for our adversarial robustness training where we consider input perturbations $\rho$ along with the the examples in clean dataset $D$. To simplify the data augmentation process, we specifically consider perturbations only on the saliency map of input data, i.e., where the corresponding saliency score \(S_i = \left\| \frac{\partial f(X)}{\partial x_i} \right\| \) is beyond a threshold. \( y' = h_\theta(X') \) is the output of the prediction model under adversarial input, and \( I(\rho; y') \) represents MI between \( \rho \) and \( y' \). The parameter \( \lambda \) balances the adversarial loss and MI under adversarial perturbations. 


To show why training to minimize MI increases the robustness, we define the adversarial risk \( R_{\text{adv}}(\theta) \) as
\begin{equation}
    R_{\text{adv}}(\theta) = \mathbb{E}_{(X, y)} \left[ \mathbb{P}\left( h_\theta(X + \rho(X)) \neq y \right) \right]
\end{equation}
Since $y' = h_\theta(X + \rho)$, both $\rho$ and $y'$ are functions of $X$. The MI term, $I(\rho; y')$, quantifies the dependency between $\rho$ and $y'$ induced by $X$. Pinsker's inequality relates the total variation distance $\delta$ between two distributions to their Kullback-Leibler (KL) divergence \cite{fedotov2003refinements}:
\begin{equation}
    \delta = \frac{1}{2} \| P_{y'|\rho} - P_{y'} \|_1 \leq \sqrt{\frac{1}{2} D_{\text{KL}}(P_{y'|\rho} \| P_{y'})}.
\end{equation}
Since $\mathbb{E}_{\rho}[D_{\text{KL}}(P_{y'|\rho} \| P_{y'})] = I(\rho; y')$, following Jensen's inequality, we get \cite{mcshane1937jensen}:
\begin{equation}
\mathbb{E}_{\rho}[\delta] \leq \sqrt{\frac{1}{2} I(\rho; y')}.
\end{equation}
To relate the adversarial risk to the total variance, we define ${P_e(\rho)}$ as the Error Probability given ${\rho}$:
\begin{equation}
P_e(\rho) = \mathbb{P}\left( h_\theta(X + \rho) \neq y \right).
\end{equation}
The change in error probability due to $\rho$ is \(\Delta P_e = P_e(\rho) - P_e,\) where $P_e = \mathbb{P}\left( h_\theta(X) \neq y \right)$ is the standard error probability without perturbations. Total variation distance is defined as: 
\begin{equation}
\Delta P_e = \left| \mathbb{P}\left( h_\theta(X + \rho) \neq y \right) - \mathbb{P}\left( h_\theta(X) \neq y \right) \right| \leq \delta.
\end{equation}
Therefore, we have \(\mathbb{E}_{\rho}[\Delta P_e] \leq \sqrt{\frac{1}{2} I(\rho; y')}.\) Since ${\Delta P_e}$ is bounded by ${\sqrt{\frac{1}{2} I(\rho; y')}}$, minimizing ${I(\rho; y')}$ will minimize ${\Delta P_e}$. Hence, reducing $I(\rho; y')$ leads to a smaller increase in error probability due to adversarial perturbations and training to minimize ${I(\rho; y')}$ lowers adversarial risk \( R_{\text{adv}} \). By minimizing this MI, the model effectively suppress its over-reliance on specific input-output correlations.

\subsection{Neural Extraction of Natural and Adversarial MI}
A critical complexity of our framework is that estimating MI itself is challenging due to the high dimensionality and non-linearity of both the input data and the network's internal representations. Neural networks map inputs to predictions through multiple layers of transformations, complicating the direct computation of the joint probability distributions necessary for MI. To extract MI from adversarial examples, we build upon Mutual Information Neural Estimation (MINE) \cite{belghazi2018mine}. MI between the input and the output of target model is given as:
\begin{equation}
I(X'; y') = \mathbb{E}_{p(X', y')}\left[\log \frac{p(X', y')}{p(X')p(y')}\right],
\end{equation}
MINE approximates this via a neural network \(T_\phi\) and optimizes it using gradient descent:
\begin{align}
\hat{I}(X'; y') &= \mathbb{E}_{p(X', y')}[T_\phi(X', y')] \nonumber\\
&\quad - \log\left(\mathbb{E}_{p(X')p(y')}\left[e^{T_\phi(X', y')}\right]\right).
\end{align}
Given \(X' = X + \rho\), to disentangle MI into natural MI (\(I_N\)) and adversarial MI (\(I_A\)), using Theorem 1 from \cite{zhou2022improving}, the MI can also be expressed as:
\begin{align}
I(X'; y') &= I(X; y') + I(\rho; y') - I(X; \rho; y') \nonumber\\
&\quad + H(y' \mid X, \rho) - H(y' \mid X').
\end{align}
Under the assumptions that \(I(X; \rho; y')\) and \(H(y' \mid X, \rho) - H(y' \mid X')\) are small, we approximate:
\begin{equation}
I(X'; y') \approx I(X; y') + I(\rho; y'). 
\end{equation}
Here, \(I(X; y')\) corresponds to natural MI (\(I_N\)) and \(I(\rho; y')\) corresponds to adversarial MI (\(I_A\)). Thus, the disentangled MIs are \(I_N := I(X; y')\) and \(I_A := I(\rho; y')\).

\begin{figure}
    \centering
    \includegraphics[width=\linewidth]{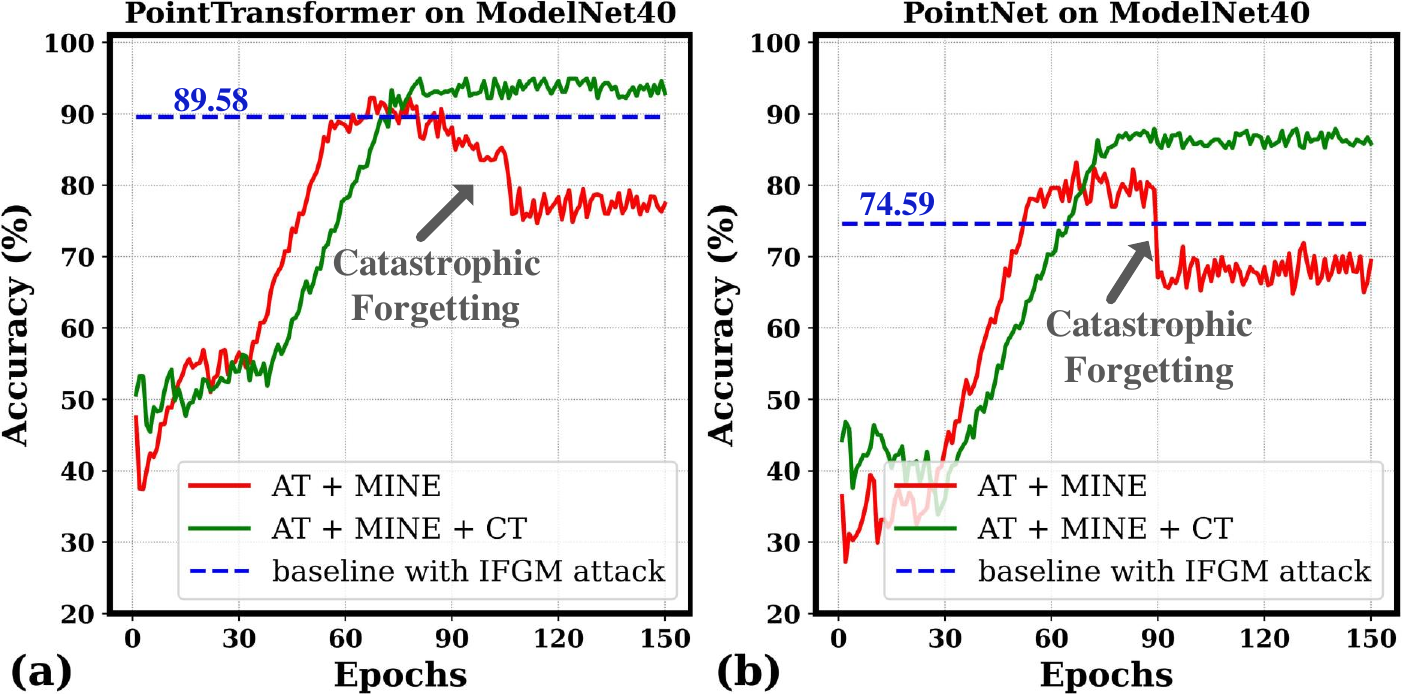}
    \caption{Directly training with the proposed MI-based objective leads to catastrophic forgetting. See AT + MINE results in (a) for PointTransformer and (b) for PointNet on ModelNet40. This is due to the conflicting goals of minimizing the prediction loss, which leverages gradient sensitivity, and minimizing the adversarial loss, which reduces sensitivity to adversarial perturbations. We address this by integrating curriculum training with the proposed MI-based approach (AT + MINE + CT results).}
    \label{fig:forget}
\end{figure}
Finally, to estimate \(I_N\) and \(I_A\), we use two separate MINE networks, \(T_{\phi_N}\) and \(T_{\phi_A}\), where the natural MI estimator is trained on natural instances \(X\) to estimate \(I_N\), while the adversarial MI estimator is trained on adversarial instances \(X' = X + \rho\) to estimate \(I_A\). The training objectives for the MINE estimators are modified as follows:
\begin{align}
\hat{I}_N &= \arg\max_{\phi_N}  \hat{I}(X; y') - \mathbb{E}_{p(X', y')}[T_{\phi_N}(X', y')] \\
\hat{I}_A &= \arg\max_{\phi_A} \hat{I}(X'; y') - \mathbb{E}_{p(X, y')}[T_{\phi_A}(X, y')]
\end{align}
In the above training framework, positive samples pair input data (natural or adversarial) with corresponding output logits, while negative samples pair the same input with shuffled logits. MINE networks are updated via gradient descent to maximize MI for positive samples and minimize it for negative ones. 

\section{Curriculum Training to Balance Prediction and Adversarial Losses}
Fig. \ref{fig:forget} illustrates the performance of MI-based adversarial training on ModelNet40 \cite{wu20153d} for PointNet and PointTransformer. The results for MINE-based adversarial training (AT + MINE) are shown in red. Notably, the previous framework leads to catastrophic forgetting. This occurs due to the conflicting objectives: minimizing prediction loss, which relies on gradient sensitivity to align the model's predictions with the ground truth, whereas minimizing input-output MI reduces gradient sensitivity to perturbations.

To address this, we integrated curriculum training (CT) with the above framework. CT, introduced by Bengio et al. \cite{bengio2009curriculum}, mimics the human learning process: starting with easier concepts and gradually progressing to more complex ones to ensure that the model is neither overwhelmed by challenging examples nor constrained by overly simple ones. 

\begin{figure}
    \centering
    \includegraphics[width=0.99\linewidth]{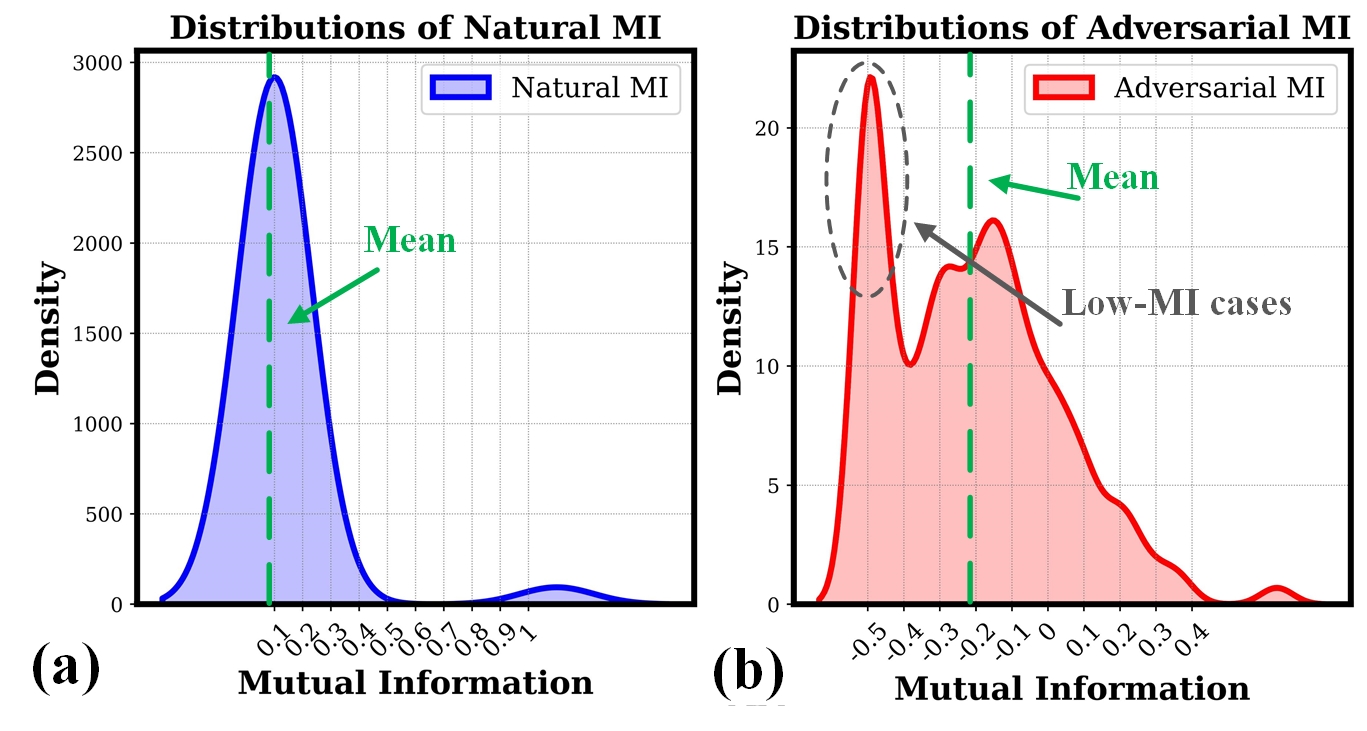}
    \caption{MI distribution for (a) natural and (b) adversarial perturbations on ModelNet40 with PointNet-based predictions. The MI for natural data clusters around the mean with a tail toward lower values, while adversarial data shows a skewed distribution, peaking at lower MI values and extending higher based on perturbation strength.}
    \label{fig:mi}
\end{figure}

\begin{figure*}[t!]
    \centering
    \includegraphics[width=0.99\linewidth]{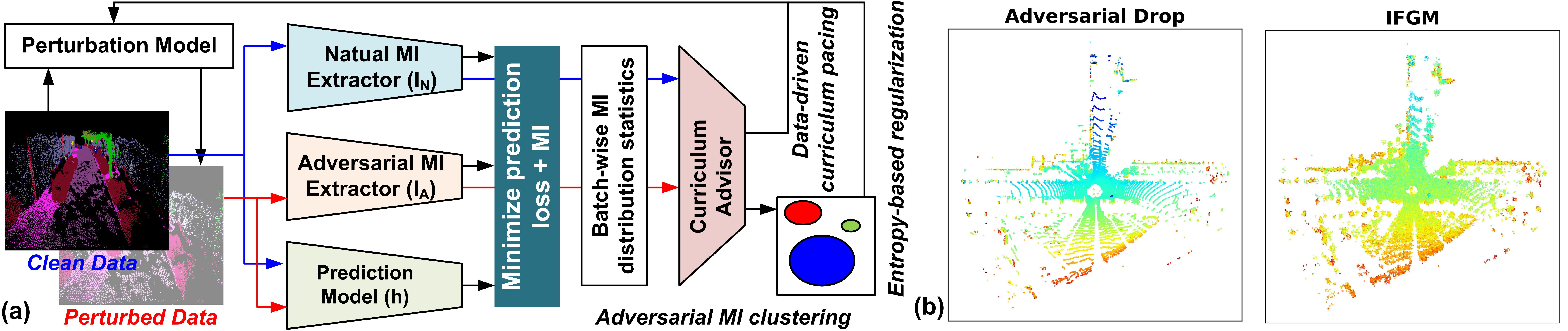}
\caption{\textbf{Overall training framework}: (a) A parametric perturbation generates perturbed data from clean data. Mutual information is extracted for both natural and adversarial data, while the prediction model receives both inputs. Optimization focuses on minimizing prediction loss and MI. After batch-wise training, summary statistics of natural and adversarial MI are sent to the curriculum advisor, which disentangles adversarial cases, clusters them, and generates entropy-based regularization. The curriculum advisor's parameters are learned through a closed-loop process, following the loss in Eq. 19. (b) Representative profiles of LiDAR point cloud manipulations under adversarial drop and IFGM that were studied to characterize the framework.}
    \label{fig:overall}
\end{figure*}

\subsection{Adversarial Training Curriculum Advisors}
We employed a neural network-based \textit{curriculum advisor} to dynamically guide the pacing of adversarial training. The advisor outputs a pacing parameter \( \eta(t) \), which adjusts the difficulty of training examples over time. Initially, MI regularization is relaxed to establish a solid foundation, with a gradual shift toward stronger MI regularization as training progresses.

To differentiate between natural and adversarial perturations, Fig. \ref{fig:mi} compares the MI of natural and adversarial data, extracted on ModelNet40 using neural networks \( \hat{I}_N \) and \( \hat{I}_A \), respectively. The MI for natural data clusters around a central value (mean MI) with a tail extending toward lower values. In contrast, the MI distribution for adversarial data is skewed, peaking at lower MI values due to adversarial noise, with the tail extending toward higher values based on the strength of perturbations. Given these trends, we integrated a dynamic strategy into the curriculum advisor to adjust the training objective based on real-time MI distribution analysis by providing it the summary statistics of MI distributions such as mean, variance, and skewness.

Additionally, as the curriculum progresses, the increased focus on low-MI adversarial perturbations can cause overfitting, leading to catastrophic forgetting. To counter this, curriculum advisors also cluster MI values into low, medium, and high using k-means and track the frequency of low-MI cases. We introduce an entropy-based regularizer to promote diversity:
\begin{equation}
H(I_A) = -\sum_{i} p(I_{A,i}) \log p(I_{A,i}),
\end{equation}
where \( p(I_{A,i}) \) represents the MI value distribution. The entropy \( H(I_A) \) quantifies diversity in adversarial MI. To adjust the loss function, we define an adaptive function that accounts for both low-MI frequency and entropy:
\begin{equation}
\Lambda(f_{\text{low}}, H(I_A)) = \alpha (1 - f_{\text{low}}) \exp(-\beta H(I_A)),
\end{equation}
where \( f_{\text{low}} \) is the frequency of low-MI cases, and \( \alpha \) and \( \beta \) are hyperparameters. 

In summary, the curriculum advisor \( P_\psi \), generates pacing function $\eta(t)$ based on mean MI (\( \mu_{I_N}, \mu_{I_A} \)), variance (\( \sigma^2_{I_N}, \sigma^2_{I_A} \)), skewness (\( \text{skew}(I_N), \text{skew}(I_A) \)), frequency of low-MI cases (\( f_{\text{low}} \)) and entropy (\( H(I_N), H(I_A) \)), i.e., 
\begin{align}
    \eta(t) &= P_\psi(\mu_{I_N},\ \mu_{I_A},\ \sigma^{2}_{I_N},\ \sigma^{2}_{I_A},\ \text{skew}(I_N), \nonumber \\
    &\quad \text{skew}(I_A),\ f_{\text{low}},\ H(I_N),\ H(I_A))
\end{align}



\subsection{Integrated Loss and Overall Framework}
We integrate both adversarial MI minimization and curriculum training components into a comprehensive loss function that combines standard adversarial training with MI and entropy-based regularization:
\begin{align}
L_{\text{total}} &= \min_{\theta, \psi} \ \mathbb{E}_{(X, y)\sim D} \Big[ L\left( h_\theta(X), y \right) \nonumber \\
&\quad +\eta(t) \cdot \Big( \max_{\rho \in \mathcal{S}} L\left( h_\theta(X'), y \right) + \lambda \cdot I(\rho; y') \Big) \Big] \nonumber \\
&\quad + \Lambda(f_{\text{low}}, H(I_A)) \cdot \gamma(f_{\text{low}}),
\end{align}
where \( \theta \) are the model parameters, and \( \psi \) are the parameters of neural network curriculum advisor \( P_\psi \) that learns the optimal pacing function \( \eta(t) \). Here, \( \eta(t) \in [0, 1] \) is a pacing parameter that adjusts the emphasis on adversarial examples over time \( t \). 

In training with the above objective, the MI term \( I(\rho; y') \) limits the predictability of adversarial perturbations on the model output, reducing adversarial risk. The entropy-based regularization \( \Lambda(f_{\text{low}}, H(I_A)) \cdot \gamma(f_{\text{low}}) \) promotes diversity, preventing overfitting and enhancing generalization. The adaptive pacing function \( \eta(t) \) ensures a balance between overly challenging examples and simpler cases. Fig. \ref{fig:overall}(a) summarizes the overall training framework and Fig. \ref{fig:overall}(b) demonstrates the representative attacks that were used to rigorously characterize the framework.



\begin{table*}
    \centering
\caption{Classification accuracy (\%) of five point cloud models (PointNet \cite{qi2017pointnet}, PointNet++ \cite{qi2017pointnet++}, DGCNN \cite{wang2019dynamic}, PointConv \cite{wu2019pointconv}, and PointTransformer \cite{zhao2021point}) on the ModelNet40 \cite{wu20153d} dataset under various adversarial attacks. Results compare performance with no defense, hybrid training \cite{ji2023benchmarking}, and our proposed method. Our method demonstrates superior robustness, particularly against Add, Drop, Perturb, IFGM, and AdvPC attacks. Bold indicates the best result for each model, and underline indicates the best overall result for each attack.}
    \begin{adjustbox}{width=2\columnwidth,center}
    \begin{tabular}{c c | c c c c c c c c c c c c }
        \hline
        \textbf{Defense} & \textbf{Model} & \textbf{Clean} & \textbf{ADD} & \textbf{Drop} & \textbf{Perturb} & \textbf{IFGM} & \textbf{PGD} & \textbf{KNN}& \textbf{GeoA3} & \textbf{L3A} & \textbf{AdvPC} & \textbf{SIA} & \textbf{AOF}\\
        \hline
         & PointNet & 87.64 & 71.76 & 59.64 &  85.58  & 74.59 & 34.32 & 45.1 & 61.26 & 45.38 & 76.94 & 31.4 & 54.54 \\
        
         & PointNet++ & 89.3 & 72.37 & 71.74 &  88.17 & 81.22 &15.54 & 54.25 & 74.51 & 44.89 & 73.62 & 16.82 & 60.01\\
        
        \textbf{No Defense} & DGCNN & 89.38 & 83.71 & 73.1 & 88.74 & 86.91 & 18.96 & 70.1 & 77.39 & 57.25 & 76.86 & 51 & 62.84 \\
        
         & Pointconv & 88.65 & 85.15 & 76.5& 88.09 & 86.51 & 9.81 & 71.8 & 77.67 & 47.57 & 76.9& 25.4 & 51.26  \\
        
         & PointTransformer & \textbf{\underline{93.8}} & 87.42 & 76.57 & 93.07& 89.58 & 27.91 & 78.3 & 83.26 & 65.1 & 81.77 & 63.38 & 69.07  \\
        
        \hline
        \hline
        
         & PointNet & 88.57 & 83.39 &  77.55 & 87.64 & 85.17 & 80.15 & 64.71 & 75.45 & \textbf{50.28} & 84.12& 53.08 & 73.34 \\
         
         & PointNet++ & 89.75 & 85.45 & 85.74 & 89 & 88.29 & 77.39 &  74.68& \textbf{82.74} & \textbf{57.74} & 86.08& 52.8& 79.85 \\
         
        \textbf{Hybrid Training \cite{ji2023benchmarking}} & DGCNN & 89.47 & 87.76 & 86.14 & 89.95 & 88.53 & 81.4& 80.59 & 82.74 & 61.14 & 87.28 & 66.29& 80.19\\
        
         & Pointconv & \textbf{90.19} & 88.82 & 86.47 &  90.28 & 89.83 & 80.06 & 81.69 & 83.31 & 57.37 & 85.13 & 45.3 & 69.65\\
         
         & PointTransformer & 93.71 & 90.58 & 89.31 & 92.86 & 92.82 & 85.15 & 85.33 & 86.9 & 63.71 & 88.05 & 71.22 & 84.59   \\

        \hline
        \hline
         & PointNet & \textbf{88.73} & \textbf{87.17} & \textbf{84.82} & \textbf{89.91}& \textbf{87.36} & \textbf{84.71} & \textbf{69.34} & \textbf{78.54} & 48.83 &\textbf{87.16} & \textbf{55.57} & \textbf{76.65}  \\
         
         & PointNet++ & \textbf{90.51} & \textbf{88.71} & \textbf{86.13} & \textbf{90.78} & \textbf{90.74} & \textbf{85.14} & \textbf{75.26} & 82.61 & 56. 35 & \textbf{88.89} & \textbf{57.05} & \textbf{81.6}\\
         
        \textbf{Ours} & DGCNN & \textbf{90.75} & \textbf{89.01} & \textbf{87.41} & \textbf{91.1} & \textbf{90.54} & \textbf{84.72} &\textbf{82.58} & \textbf{83.95} & \textbf{66.4} & \textbf{89.07} & \textbf{70.52} & \textbf{83.49} \\
        
         & Pointconv & 88.89 & \textbf{89.25} & \textbf{87.93} & \textbf{90.63} & \textbf{90.7} & \textbf{82.73} & \textbf{82.51} & \textbf{84.12} & \textbf{65.38} & \textbf{88.78} & \textbf{53.27} & \textbf{73.62}     \\
         
         & PointTransformer & 93.63 & \textbf{\underline{92.94}} & \textbf{\underline{90.52}} & \textbf{\underline{93.55}} & \textbf{\underline{94.07}} & \textbf{\underline{89.11}} & \textbf{\underline{88.43}} & \textbf{\underline{90.95}} & \textbf{\underline{69.89}} & \textbf{\underline{90.7}} & \textbf{\underline{73.27}} & \textbf{\underline{87.75}}   \\

        \hline
    \end{tabular}
    \end{adjustbox}
    \label{tab:modelnet40_table}
\end{table*}

\begin{table*}
    \centering
\caption{Object detection mean Average Precision (mAP) of three models (PointPillars \cite{lang2019pointpillars}, SECOND \cite{yan2018second}, and TED \cite{wu2023transformation}) on the KITTI dataset \cite{Geiger2012CVPR} under various adversarial attacks. The results compare performance with no defense and our proposed method. Our method demonstrates superior robustness across most attacks. Bold indicates the best result for each model, and underline indicates the best overall result for each attack.}
    \begin{adjustbox}{width=2\columnwidth,center}
    \begin{tabular}{c c | c c c|| c c c|| c c c|| c c c|| c c c }
        \hline
        \textbf{Defense} & \textbf{Model} & \multicolumn{3}{c||}{\textbf{Clean}} & \multicolumn{3}{c||}{\textbf{ADD}} & \multicolumn{3}{c||}{\textbf{Drop}} & \multicolumn{3}{c||}{\textbf{Perturb}} & \multicolumn{3}{c}{\textbf{IFGM}}\\

        &  & Car & Ped &Cyclist& Car & Ped & Cyclist&Car &Ped& Cyclist & Car &Ped& Cyclist & Car & Ped & Cyclist \\
        \hline
        
         & PointPillar & 77.31 & 52.33 & 62.78& 66.5 & 40.92 & 52.98 & 56.61 &34.42 & 51.63 & 75.13 & 49.66 & 60.73 & 51.71 & 30.59 & 48.83 \\
         
        \textbf{No Defense} & SECOND & 78.63 & 53 & 67.17& 68.75 & 42.83 &54.73 & 65.37 & 37.18 & 54.22 & 75.9 &51.17 &60.93 & 50.97 & 35.04& 50.19  \\
        
         & TED & 87.91 & \textbf{\underline{67.86}} & \textbf{\underline{75.83}} & 78.57 &54.15 & 61.92 & 81.36 & 62.12 & 68.37 & 84.92 &62.69 & 71.35 & 76.14 & 55.83 & 64.86  \\
        \hline
        \hline

        & PointPillars & \textbf{78.78}& \textbf{52.73} & \textbf{63.65}& \textbf{76.91} & \textbf{50.74} & \textbf{61.85} & \textbf{74.11} & \textbf{47.43} & \textbf{58.91} & \textbf{77.21} & \textbf{52.4} & \textbf{62.75} & \textbf{76.96} & \textbf{51.37} & \textbf{62.88} \\
        
        \textbf{Ours} & SECOND & \textbf{79.37} & \textbf{53.38} & \textbf{68.47} & \textbf{78.52} & \textbf{52.69} & \textbf{65.95} &\textbf{74.92} & \textbf{48.84} &\textbf{64.75} & \textbf{78.71} & \textbf{52.89} & \textbf{67.5} & \textbf{77.81} & \textbf{53.07} & \textbf{67.39 } \\
        
         & TED & \textbf{\underline{88.29}} & 67.02 & 75.74  & \textbf{\underline{87.13}} & \textbf{\underline{65.7}} & \textbf{\underline{73.83}} & \textbf{\underline{86.03}} &\textbf{\underline{64.27}} & \textbf{\underline{71.3}} & \textbf{\underline{87.76}} & \textbf{\underline{66.41}} & \textbf{\underline{74.4}} & \textbf{\underline{87.15}} & \textbf{\underline{66.27}} & \textbf{\underline{73.29}}  \\
        
        \hline
    \end{tabular}
    \end{adjustbox}
    \label{tab:kitti_table}
\end{table*}

\begin{table*}
    \centering
    \caption{\textbf{Ablation Study:} showing the impact of adversarial training (AT), low adversarial MI (MINE), and data-driven curriculum training (CT) on the detection accuracy (mAP) of PointPillars \cite {lang2019pointpillars} and TED \cite{wu2023transformation} models on the KITTI dataset \cite{Geiger2012CVPR} across Car, Pedestrian, and Cyclist classes under clean and adversarial attack scenarios (ADD, Drop, Perturb, IFGM).}
    \begin{adjustbox}{width=2\columnwidth,center}
    \begin{tabular}{ c || c c c  || c c c|| c c c||c c c|| c c c|| c c c }
        \hline
        \textbf{Model} & \multicolumn{3}{c||}{\textbf{Defense Setup}} & \multicolumn{3}{c||}{\textbf{Clean}} & \multicolumn{3}{c||}{\textbf{ADD}}& \multicolumn{3}{c||}{\textbf{Drop}} & \multicolumn{3}{c||}{\textbf{Perturb}} & \multicolumn{3}{c}{\textbf{IFGM}}\\

        & \textbf{AT} & \textbf{MINE} & \textbf{CT} &  Car & Ped &Cyclist& Car & Ped & Cyclist&Car &Ped& Cyclist & Car & Ped &Cyclist & Car & Ped &Cyclist \\
        \hline
        
        \multirow{4}{5em}{\textbf{PointPillars}} & \xmark& \xmark &\xmark & 77.31 & 52.33 & 62.78& 66.5 & 40.92 & 52.98 & 56.61 &34.42 & 51.63 & 75.13 & 49.66 & 60.73 & 51.71 & 30.59 & 48.83 \\
        
        & \cmark & \xmark &\xmark & 79.02 & 52.89 & \textbf{63.71} & 68.84 & 42.7  & 54.32 & 60.64 & 38.21 & 52.99 & 75.49 & 50.05 & 60.52 &59.53 & 37.6 & 52.13 \\
        
        &\cmark & \cmark &\xmark & \textbf{79.12} & \textbf{53} & 63.29 & 70.55 & 45.38 & 57.23  & 65.07 & 42.92 & 54.37 & 75.81 & 50.89 & 61.03 & 64.5 & 43.41& 55.73\\

         & \cmark  &\cmark & \cmark & 78.78& 52.73 & 63.65& \textbf{76.91} & \textbf{50.74} & \textbf{61.85} & \textbf{74.11} & \textbf{47.43} & \textbf{58.91} & \textbf{77.21} & \textbf{52.4} & \textbf{62.75} & \textbf{76.96} & \textbf{51.37} & \textbf{62.88} \\
         \hline
         \hline

         \multirow{4}{5em}{\textbf{TED}} & \xmark & \xmark & \xmark & 87.91 & 67.86 & 75.83 & 78.57 &54.15 & 61.92 & 81.36 & 62.12 & 68.37 & 84.92 &62.69 & 71.35 & 76.14 & 55.83 & 64.86 \\
         
        & \cmark & \xmark &\xmark & 88 & 67.53& 75.81& 80.13 & 56.74  & 63.73 & 83.29 & 62.84 & 69.5 & 85.11 &63.53 &72.49 &78.38& 57.02 & 66.27\\
        
        &\cmark & \cmark &\xmark & 88.21& \textbf{\underline{67.88}}& \textbf{\underline{76.03}} & 84.23 & 60.15 & 67.91 & 84.97 & 63.05 & 70.71& 86.19 &64.82 &72.97& 82.86 & 60.9 & 68.88 \\

         & \cmark  &\cmark & \cmark &  \textbf{\underline{88.29}} & 67.02 & 75.74  & \textbf{\underline{87.13}} & \textbf{\underline{65.7}} & \textbf{\underline{73.83}} & \textbf{\underline{86.03}} &\textbf{\underline{64.27}} & \textbf{\underline{71.3}} & \textbf{\underline{87.76}} & \textbf{\underline{66.41}} & \textbf{\underline{74.4}} & \textbf{\underline{87.15}} & \textbf{\underline{66.27}} & \textbf{\underline{73.29}} \\

        \hline
    \end{tabular}
    \end{adjustbox}
    
    \label{tab:ablation study}
\end{table*}

\section{Simulation Results}
\subsection{Classification Results}
Table~\ref{tab:modelnet40_table} shows the classification accuracy of five state-of-the-art point cloud frameworks—PointNet \cite{qi2017pointnet}, PointNet++ \cite{qi2017pointnet++}, DGCNN \cite{wang2019dynamic}, PointConv \cite{wu2019pointconv}, and PointTransformer \cite{zhao2021point}—on the ModelNet40 dataset \cite{wu20153d} under various adversarial attacks. Our defense mechanism significantly outperforms the hybrid training method from \cite{ji2023benchmarking}, which already surpassed other state-of-the-art defenses like DUP-Net \cite{zhou2019dup}, IF-Defence \cite{wu2020if}, and SOR \cite{zhou2019dup} by $2-5\%$. Therefore, accuracy improvements in the proposed framework are remarkable. Moreover, our method's accuracy improvements are consistent across attack strategies such as Add, Drop, Perturb, IFGM, and AdvPC, though it struggles with more spatially-aware attacks like KNN and L3A. Additionally, PointTransformer combined with our training method outperforms other frameworks across most attack types. For instance, the PointTransformer maintains high accuracy under GeoA3, L3A, and AOF attacks, where other models show significant performance degradation. This aligns with transformers' strengths in capturing spatial relationships and interactions between points, which are critical for understanding point cloud geometry. 

\subsection{Object Detection Results}
Table \ref{tab:kitti_table} presents the performance of three leading 3D object detection frameworks—PointPillars \cite{lang2019pointpillars}, SECOND \cite{yan2018second}, and TED \cite{wu2023transformation}—on the KITTI dataset \cite{Geiger2012CVPR} under various adversarial attacks, targeting key object classes: Car, Pedestrian, and Cyclist. Our defense method consistently outperformed the no-defense scenario, with TED showing substantial gains, particularly for the Car and Cyclist classes. For instance, TED’s mAP for the Car class under the ADD attack improved from 78.57\% to 87.13\%, and from 76.14\% to 87.15\% under the IFGM attack. Simpler models like PointPillars and SECOND also benefited, with SECOND’s mAP for the Car class under the Drop attack rising from 65.37\% to 74.92\%. These results highlight our method's broad improvement in robustness.

Table~\ref{tab:ablation study} presents an ablation study evaluating the impact of different components of our defense mechanism—adversarial training (AT), adversarial training with mutual information minimization, and curriculum training (CT)—on the performance of PointPillars \cite{lang2019pointpillars} and TED \cite{wu2023transformation}. AT alone shows modest improvements, but when combined with MI minimization, the improvements are pronounced. For instance, in the TED model, Cyclist class mAP under the IFGM attack increases from 64.86\% to 68.88\%, and in PointPillars, Pedestrian class mAP rises from 37.6\% to 43.41\%. However, as in Fig.~\ref{fig:forget}, the integrated framework is prone to catastrophic forgetting.

The combination of adversarial training, low-mutual information cases, and curriculum training (AT + MINE + CT) yields the best overall performance in the table. In the TED model, this setup boosts Car detection accuracy under the ADD attack from 84.23\% to 87.13\% and from 83.29\% to 86.03\% under the Drop attack. Curriculum training gradually emphasizes harder, low-MI cases, improving robustness to a wider range of attacks. For instance, TED's accuracy under IFGM for Pedestrian rises from 55.83\% to 66.27\%, showing CT's effectiveness against complex attacks. The substantial gains, especially for challenging classes like Cyclist and Pedestrian, highlights that combining all three components is critical for comprehensive defense.

\section{Conclusion}
Our curriculum training framework, which minimizes adversarial mutual information (MI), enhances 3D vision models' robustness, outperforming state-of-the-art methods. It achieves 2–5\% accuracy gains on ModelNet40 and a 5–10\% mAP boost on KITTI. In the TED model, our defense improved Car detection under ADD from 78.57\% to 87.13\%, and resilience against IFGM from 76.14\% to 87.15\%. Ablation studies confirm that combining adversarial training (AT), MI extraction (MINE), and curriculum training (CT) provides the best results, significantly improving performance across challenging classes and attacks.

\newpage
\bibliographystyle{IEEEtran}
\bibliography{main}

\end{document}